\documentclass[lettersize,journal]{IEEEtran}
\usepackage{amsmath,amsfonts}
\usepackage{algorithmic}
\usepackage{algorithm}
\usepackage{array}
\usepackage[caption=false,font=normalsize,labelfont=sf,textfont=sf]{subfig}
\usepackage{textcomp}
\usepackage{stfloats}
\usepackage{url}
\usepackage{verbatim}
\usepackage{graphicx}
\usepackage{cite}
\usepackage{amsthm}
\newtheorem{definition}{Definition} 
\newtheorem{lemma}{Lemma} 

\usepackage{booktabs}
\usepackage{multirow}
\newcommand{\tabincell}[2]{\begin{tabular}{@{}#1@{}}#2\end{tabular}}

\begin{document}

\title{SGPC: Safety-Guiding Policy Constraint\\for Stable Off-Policy Safe Reinforcement Learning}

\author{Yifan Liu,~\IEEEmembership{Member, IEEE}, Nan Zhang, Yang Cong,~\IEEEmembership{Senior Member, IEEE}
\thanks{Yifan Liu and Yang Cong are with the School of Automation Science and Engineering, South China University of Technology, Guangzhou, 510641, China (e-mail: yifanliu@scut.edu.cn, congyang81@gmail.com).}
\thanks{Nan Zhang is with the China Southern Power Dispatching \& Control Center, China Southern Power Grid Company Limited, Guangzhou, 510641, China (e-mail: zhangnan1116@csg.cn).}
}

\markboth{Journal of \LaTeX\ Class Files,~Vol.~14, No.~8, August~2021}%
{Shell \MakeLowercase{\textit{et al.}}: A Sample Article Using IEEEtran.cls for IEEE Journals}

\IEEEpubid{0000--0000/00\$00.00~\copyright~2021 IEEE}

\maketitle

\begin{abstract}
Off-policy safe reinforcement learning (RL) faces a challenging trade-off between task optimality and constraint feasibility. In primal-dual methods, the coupled optimization of cumulative reward and constraint penalties through adaptive Lagrange multipliers may induce oscillatory policy updates near the safety boundary and amplify the effect of unreliable reward estimates outside the safe set. To address these issues, we formulate a new constrained optimization problem (COP) for safe RL by introducing a Safety-Guiding Policy (SGP) and a corresponding Safety-Guiding Policy Constraint (SGPC). The proposed COP augments the standard COP in safe RL with a conditional safety-guidance constraint. During actor updates, the current actor is evaluated using the learned cost critic. If it is predicted to be outside the safe set, SGP provides a nearby lower-cost training reference, and SGPC guides actor optimization toward this reference. Once the policy enters the safe set, the added constraint becomes inactive, and the formulation reduces to the original COP. This formulation leads to a plug-in actor-training constraint that can be integrated into off-policy safe RL algorithms. By applying primal-dual optimization to the auxiliary COP, we finally derive an SGPC-based off-policy safe RL algorithm and implement it in two representative off-policy algorithms, yielding SAC-SGPC and TD3-SGPC. Experiments on multiple Safety-Gymnasium benchmarks demonstrate that SGPC achieves a better optimality-safety trade-off by substantially reducing constraint violations and cost oscillations while maintaining competitive or better task returns.
\end{abstract}


\section{Introduction}

\IEEEPARstart{I}{n} the past few years, reinforcement learning (RL) has been extensively studied and applied in a wide range of domains, such as robotics and power systems \cite{RL_review_robot,RL_robot_Egli_2024,RL_robot_luo_2025,RL_robot_yuntao_2025,RL_power_cao_2025,app_safe_2025}. These approaches typically employ neural networks to solve an unconstrained optimization problem by maximizing the cumulative rewards. While such formulations have achieved remarkable success, their deployment in real-world systems raises concerns regarding safety and reliability, which may lead to unsafe actions in many real-world tasks. In view of safety considerations, safe RL explores RL algorithms by formulating safety concerns as constraints. In recent years, safe RL has garnered significant attention driven by the increasing demand for constraint satisfaction \cite{Review_gu_2024,review_zhao_2023}.

Safe RL can be seen as solving a special kind of constrained optimization problem (COP) for a black-box system \cite{review_garcia_2015,constraint_saferl}. And most safe RL methods employ primal-dual methods, which transform the COP (primal) to an unconstrained form (dual) by introducing Lagrange multipliers to penalize the constraints \cite{Lag_RL-Ha_2021}. Primal-dual methods alternately and iteratively update the policy network and Lagrange multiplier, the concept of which originates from the Lagrangian method in constrained optimization theory \cite{Lag_RL_Ray_2019}. However, primal-dual methods encounter severe oscillations in the reward and cost curves during the training process, especially when the policy approaches the boundary of the safe set \cite{lyapunov-safe-rl}. Furthermore, they often struggle with the trade-off between optimality (high cumulative reward) and constraint feasibility (low cumulative cost) \cite{state-wise-safe-rl-zhan-2024,balance-gu-2024,P3O_2022}. 

\IEEEpubidadjcol

Why does this happen? We focus on two possible contributing factors. First, under the standard Lagrangian update, the Lagrange multiplier is employed to penalize constraint violations and decreases when the policy enters the safe set. The resulting reduction in the multiplier allows the reward term to dominate and may move the policy back outside the safe set, potentially resulting in increased constraint violations, producing policy oscillation around the safe set boundary and recurrent unsafe behavior \cite{CVPO}. Second, the distribution of the reward value function may differ inside and outside the safe set. The critics trained mainly outside or near the safe set boundary can be unreliable upon entering it. In addition, this issue can be particularly exacerbated for off-policy methods due to their intrinsic properties \cite{Review_gu_2024}.

Based on the above analysis, we argue that stabilizing primal-dual safe RL requires not only rapidly guiding the actor toward the predicted safe set but also retaining feasibility pressure after that region has been reached. By reducing oscillation near the boundary, this mechanism may allow the policy to spend more training time in this region and improve critic learning around the state-action distributions relevant to feasible policy optimization. These effects may help the actor discover higher-reward feasible policies and thereby improve the empirical trade-off between optimality and constraint feasibility.

Motivated by this insight, we develop SGPC-based algorithms for off-policy safe RL. Specifically, we proceed as follows. First, we define an idealized SGP which can provide a nearby reference policy that is closer to or inside the safe set. Second, we formulate an SGPC-augmented auxiliary COP and characterize its conditional relationship to the original safe RL problem. Third, we design an efficient finite-sample approximation of SGP to find an estimated better reference with the learned cost critic to support SGPC-based algorithms in practice. Finally, by applying primal-dual optimization methods to the proposed COP, we construct an SGPC actor-training surrogate for off-policy safe RL. We instantiate this surrogate in SAC-Lag and TD3-Lag, yielding two practical implementations, SAC-SGPC and TD3-SGPC \cite{SAC_2018,TD3_2018,Lag_RL_Ray_2019}.

Our contributions in this paper are outlined as follows.
\begin{itemize}
\item We develop a novel COP for safe RL by integrating the original COP with SGPC and SGP. The new COP provides a theoretical formulation for SGPC-based safe RL algorithms to proactively guide the actor's policy toward the safe set, thereby improving constraint feasibility.
\item We present the SGP and SGPC as key components of SGPC-based safe RL algorithms. We also design an efficient sampling-based approximation of SGP to support SGPC-based algorithms in practice, providing a safer and proximate reference policy output when the actor is predicted to be outside the safe set.
\item We propose an SGPC-based safe RL algorithm by incorporating SGPC as a plug-in constraint into the actor update of off-policy actor-critic safe RL. We further instantiate the proposed algorithm in SAC-Lag and TD3-Lag, yielding two practical implementations, namely SAC-SGPC and TD3-SGPC. Experimental results demonstrate that incorporating SGPC substantially reduces observed constraint violations and cost oscillations while maintaining competitive or better cumulative reward.
\end{itemize}

\section{Related Work}
In recent years, safe RL has attracted increasing attention. From an optimization perspective, safe RL can be viewed as solving a black-box constrained optimization problem. As in standard RL, safe RL methods can be broadly categorized into on-policy and off-policy methods \cite{review_zhao_2023,Review_gu_2024}. 

Most existing studies focus on on-policy safe RL. To enhance safety during training, second-order-based on-policy safe RL restricts policy updates to small steps within a trust region, such as Constrained Policy Optimization (CPO) \cite{CPO_2017}, which integrates safety constraints into the TRPO algorithm \cite{TRPO_2017}. Building on CPO, Projection-based Constrained Policy Optimization (PCPO) presents a two-stage iterative method for optimizing policies \cite{PCPO_Yang_2020}. In the second stage, the policy is projected onto the safe set by solving a projection optimization problem. Besides that, first-order on-policy safe RL enforces constraints via primal–dual formulations or barrier functions \cite{IPO-2020,CUP_2022,P3O_2022}. 

For off-policy safe RL, most methods employ primal-dual methods. Lagrangian-based safe RL methods combine standard RL methods with Lagrange multipliers to handle constraints, such as SAC-Lag and TD3-Lag \cite{Lag_RL_Ray_2019,Lag_RL-Ha_2021}. Building upon these, PID-Lagrangian-based methods replace the conventional gradient-based update of the Lagrange multipliers with a PID control–inspired update scheme, such as SAC-PID and TD3-PID \cite{PIDLag_2020}. Besides that, Constrained Variational Policy Optimization (CVPO) employs an expectation-maximization approach and a trust region-based convex optimization method to incorporate constraints during policy learning \cite{CVPO}.

There are also several studies in RL and learning to optimize (L2O) that inspire our method. Some model-based RL methods employ sampling-based planning techniques, such as Model Predictive Path Integral (MPPI), to select control actions over a predicted future horizon \cite{MPPI_2015,safedreamer_2024,tdmpc2_2024}. For L2O, some methods solve a known COP by employing a gradient-based policy correction phase and Lagrangian-based training \cite{ssl_dc3,somtp_2024,ssl_lag_dual_dc3_2023}. In addition, some methods apply model-based Control Barrier Functions (CBFs) or projection-based optimization to RL outputs to improve safety \cite{CBF_DU_2023,CBF_CHENG_2019}. Besides, critic-guided action correction is also used by SEditor~\cite{SEditor_2022}, which learns a separate policy to edit actions executed in the environment, while PCRPO~\cite{balance-gu-2024} addresses reward-cost conflicts through gradient manipulation. In our method, both SGP and SGPC operate exclusively during actor updates in off-policy safe RL. Unlike an execution-time projection or safety filter, SGP neither requires the sampled reference action to be strictly feasible nor replaces the action executed in the environment. SGP provides a training reference only when the finite-sample selection objective is improved, and SGPC uses this reference to guide actor updates, encouraging and guiding the actor to progressively learn safer policies while improving task performance.

\section{Problem Formulation}
\subsection{Safe RL}
Safe RL problems can be modeled as a Constrained Markov Decision Process (CMDP) $\mathcal{M} \sim \left(\mathcal{S}, \mathcal{A}, \mathcal{P}, {r}, {c}, \gamma, d\right)$ \cite{CMDP_Altman_1998}, where $\mathcal{S}$ is the state space, $\mathcal{A}$ is the action space, $\mathcal{P}:$ $\mathcal{S}$ $\times$ $\mathcal{A}$ $\times$ $\mathcal{S}$ $\rightarrow$ $[0, 1]$ is the transition function, $r:$ $\mathcal{S}$ $\times$ $\mathcal{A}$ $\rightarrow$ $\mathbb{R}$ is the reward function, $c:$ $\mathcal{S}$ $\times$ $\mathcal{A}$ $\rightarrow$ $\mathbb{R}$ is the cost function, $\gamma$ is the discount factor, and $d\in\mathbb{R}$ represents the safety threshold or the cost limit.
Compared with traditional RL algorithms, safe RL focuses on finding an optimal policy $\pi$ within a safe set, balancing the cumulative reward and the cost for not satisfying the safety constraints. With $\rho_\pi$ denoting the trajectory distribution generated by $\pi$, 
the expected cumulative reward and cost can be defined as 
$Q_r^{\pi}(s,a)=\mathbb{E}_{(s_t,a_t)\sim\rho_\pi}\left[ \sum_{t=0}^\infty \gamma^t\cdot r(s_t,a_t)\big|s_0=s,a_0=a\right]$
and 
$Q_c^{\pi}(s,a)=\mathbb{E}_{(s_t,a_t)\sim\rho_\pi}\left[ \sum_{t=0}^\infty \gamma^t\cdot c(s_t,a_t)\big|s_0=s,a_0=a\right]$, respectively. 
The safe RL actually tries to solve the following constrained optimization problem (COP):
\begin{align}
    \max_\pi\;\;&\mathbb{E}_{s\sim p_s,a\sim\pi}\left[Q_r^{\pi}(s,a)\right] \label{eq3.1-1}\\
    \text{s.t.}\;\;&\mathbb{E}_{s\sim p_s,a\sim\pi}\left[Q_c^{\pi}(s,a)\right]\leq d \notag
\end{align}
where $s\sim p_s$ represents the distribution of initial states.

\subsection{Primal-Dual Method}
\label{sec:3.2}
The Lagrangian method is a typical kind of primal-dual method and has been introduced to safe RL to solve the COP in \eqref{eq3.1-1}. Generally speaking, the Lagrangian method introduces a Lagrange multiplier $\lambda\geq0$, giving the following Lagrangian dual formulation of the original COP.
\begin{equation}
    \min_{\lambda\geq0}\max_\pi \mathbb{E}_{s,a}\left[Q_r^{\pi}(s,a)\right]-\lambda\left(\mathbb{E}_{s,a}\left[Q_c^{\pi}(s,a)\right]-d \right) 
    \label{eq3.2-1}
\end{equation}
Based on this Lagrangian formulation, Lagrangian method-based safe RL algorithms train the policy using the dual gradient descent method, which alternates between optimizing $\pi$ and $\lambda$. Specifically, these algorithms first train the two critic networks $Q_r^\pi(s,a)$ and $Q_c^\pi(s,a)$. The policy $\pi$ and the Lagrangian multiplier $\lambda$ are then updated iteratively by minimizing the actor loss $J_\pi$ and the multiplier's loss $J_\lambda$, respectively.

\begin{equation}
    J_\pi(\phi) = \mathbb{E}_{s\sim \mathcal{D}}\left[-Q_r^\pi(s,\pi_\phi(s))+\lambda Q_c^\pi(s,\pi_\phi(s)) \right]
    \label{eq3.2-2}
\end{equation}
\begin{equation}
   J_\lambda(\lambda) = \lambda\mathbb{E}_{s\sim \mathcal{D}}\left[\left(d- Q_c^\pi(s,\pi_\phi(s)) \right) \right]
    \label{eq3.2-3}
\end{equation}
$\mathcal{D}$ denotes a batch of states sampled from the replay buffer.
By integrating the aforementioned Lagrangian method with common RL algorithms, we can obtain a series of Lagrangian-based safe RL algorithms, such as SAC-Lag, TD3-Lag, PPO-Lag, etc. \cite{Lag_RL_Ray_2019, Lag_RL-Ha_2021}.
\begin{figure*}[!tb]
  \begin{center}
    \centerline{\includegraphics[width=1.9\columnwidth]{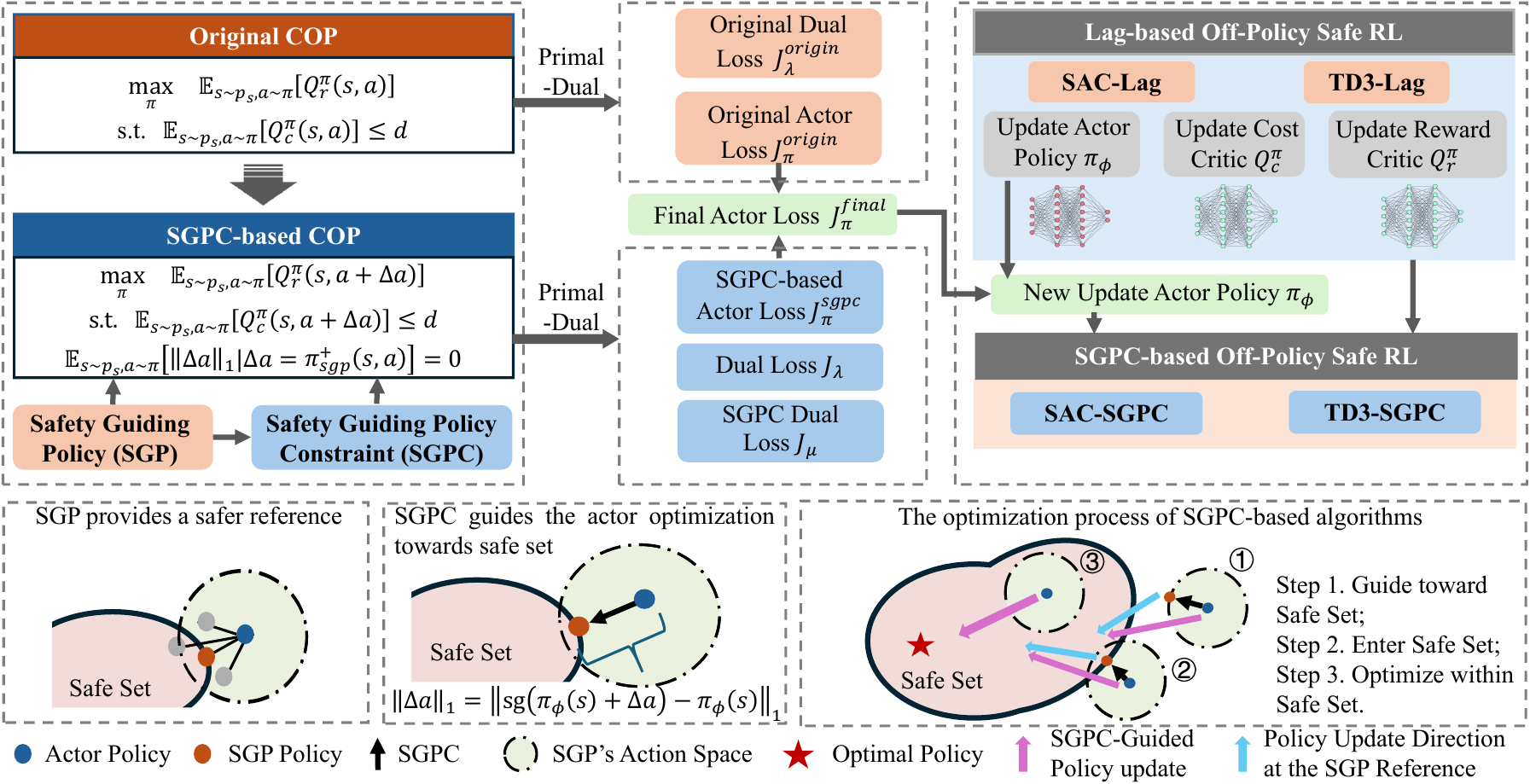}}
    \caption{Overview of the SGPC-based algorithm, which incorporates SGPC as a plug-in constraint into the actor update of off-policy actor-critic safe RL.}
    \label{sgpc_train}
  \end{center}
  \vspace{-0.5cm}
\end{figure*}
\section{Algorithm}
This section introduces the idealized SGPC formulation and its practical actor-training algorithm, as illustrated in Fig.~\ref{sgpc_train}. 
The section is structured as follows: 
In Section \ref{sec:4.1}, we construct an auxiliary COP with an additional guidance-based constraint. This new COP makes it possible to integrate the SGP into the agent's training process. In Section \ref{sec:4.2}, we present a finite-sample approximation of the SGP that searches for a locally safer reference. In Section \ref{sec:4.3}, the training algorithm with SGPC is presented. In Section \ref{sec:4.4}, we further describe how SGPC is instantiated in two off-policy safe RL algorithms, SAC-Lag and TD3-Lag, yielding SAC-SGPC and TD3-SGPC, respectively.

\subsection{A New COP with Safety-Guiding Policy Constraints}
\label{sec:4.1}
Lagrangian-based safe RL may oscillate near the constraint boundary because the cost multiplier decreases after the policy becomes feasible, allowing reward-driven updates to move it outside again. Such repeated departures might reduce the time available for learning high-reward behavior within the safe set. We therefore aim to guide the actor rapidly toward the predicted safe set during the actor update process.

We first define an ideal local policy that can make the agent's action safer in terms of the true cost critic:

\begin{definition}
  \label{def:SGP}
  \textbf{Safety-Guiding Policy (SGP).} 
  Let $v_\pi(s,a)=\operatorname{ReLU}(Q_c^\pi(s,a)-d)$. An ideal SGP $\pi_{sgp}^+:\mathcal{S}\times\mathcal{A}\rightarrow\mathbb{R}^{m}$ returns an admissible correction $\Delta a$, where $m$ is the action dimension and $a+\Delta a\in\mathcal{A}$. Given a candidate set $\mathcal{A}_{sgp}$ containing $\mathbf{0}$, and a penalty coefficient $\lambda_{sgp}>0$, it selects the $\Delta a\in\mathcal{A}_{sgp}$ minimizing $\lambda_{sgp}\|\Delta a\|_1+v_\pi(s,a+\Delta a)$. It returns a nonzero correction whenever a candidate strictly improves this safety; otherwise, it returns $\mathbf{0}$. In particular, $v_\pi(s,a)=0$ implies $\Delta a=\mathbf{0}$.
\end{definition}

When $\Delta a\neq0$, the ideal SGP has identified a nearby reference action with a lower cost violation, which can be used to guide the actor to approach the safe set. When $\Delta a=\mathbf{0}$, the original action may already be feasible, or SGP may be unable to find a sufficiently improved reference. Based on this mechanism, we define a new constraint for safe RL policies.

\begin{definition}
  \label{def:SGPC}
  \textbf{Safety-Guiding Policy Constraint (SGPC).} Given an ideal SGP $\pi_{sgp}^+$, the constraint $\mathbb{E}_{s\sim p_s,a\sim\pi}\left[\|\Delta a\|_1 \mid \Delta a= \pi_{sgp}^+(s,a)\right]=0$ is called the safety-guiding policy constraint for policy $\pi$. 
\end{definition}

The SGPC serves to guide actor training via an SGP, thereby improving constraint satisfaction. As previously stated, the objective of safe RL is to solve the COP in \eqref{eq3.1-1}. We augment this problem with SGPC and formulate the auxiliary COP in \eqref{eq4.1-1}, which formalizes the use of SGP-based guidance during policy optimization.
\begin{align}
    \max_\pi\;\;&\mathbb{E}_{s\sim p_s,a\sim\pi}\left[Q_r^{\pi}(s,a+\Delta a)\right] \label{eq4.1-1}\\
    \text{s.t.}\;\;&\mathbb{E}_{s\sim p_s,a\sim\pi}\left[Q_c^{\pi}(s,a+\Delta a)\right]\leq d \notag\\
    &\mathbb{E}_{s\sim p_s,a\sim\pi}\left[\|\Delta a\|_1 \big|\Delta a= \pi_{sgp}^+(s,a)\right]={0} \notag
\end{align}

The auxiliary COP extends the original COP in \eqref{eq3.1-1} by retaining its objective and cost constraint while introducing the additional SGPC. Their relationship is characterized as follows:

\begin{lemma}
    \label{feasible_2cop}
    If a policy $\pi$ is feasible for the COP in \eqref{eq4.1-1}, then it is also feasible for the COP in \eqref{eq3.1-1}.
\end{lemma}
\begin{lemma}
    \label{equal_2cop}
   If a policy $\pi$ satisfies the SGPC, then $\Delta a = \mathbf{0}$ holds almost surely under the induced state-action distribution. For this policy, the objective and cost expressions in \eqref{eq4.1-1} coincide with those in \eqref{eq3.1-1}.
\end{lemma}
\begin{lemma}
    \label{optimal_2cop}
    Let $\pi^*$ be the optimal solution to the original COP in \eqref{eq3.1-1}. If $\pi^*$ is feasible for the new COP in \eqref{eq4.1-1}, then $\pi^*$ is also the optimal solution for \eqref{eq4.1-1}.
\end{lemma}

Proofs are provided in \textbf{Appendix~\ref{App:B}}. Regarding Lemma~\ref{equal_2cop}, one may ask why the objective and cost constraint in \eqref{eq4.1-1} are evaluated at $a+\Delta a$ rather than at $a$ as in \eqref{eq3.1-1}. During the RL training process, SGPC may not yet be satisfied and $\Delta a$ can be nonzero. In this case, $a+\Delta a$ represents the reference action provided by SGP, and $Q_r^\pi(s,a+\Delta a)$ and $Q_c^\pi(s,a+\Delta a)$ evaluate the efficacy of the safer policy guided by the SGP. 

Furthermore, the SGPC can offer an additional training direction for actor updates using states sampled from the replay buffer. When SGP produces a nonzero guiding action for a sampled state, it has identified a safer reference for the actor's output. This suggests that the policy has not yet remained consistently within the predicted safe set over the sampled state distribution and may still produce unsafe outputs. SGPC then guides the actor toward the safe set. When SGP returns $\Delta a=\mathbf{0}$, either the current actor output is predicted to be safe or no sufficiently better reference has been found. In either case, the SGPC becomes inactive, avoiding potentially unreliable guidance. If the actor output is predicted to be safe, the actor can continue improving its reward performance within the safe set, while the retained SGPC multiplier allows the guidance to reactivate whenever SGP finds a safer guiding action, indicating that the actor has not remained stably feasible over the sampled states. Motivated by the auxiliary COP in \eqref{eq4.1-1}, we develop the corresponding practical actor-update method in Section~\ref{sec:4.3}.

\subsection{Sampling-based Safety-Guiding Policy}
\label{sec:4.2}
We now describe a practical approximation to SGP. Before that, it should be noted that the SGP is not a learned network. During actor updates, this SGP constructs a reference action for each sampled state using the current actor output and the learned cost critic.

If the feasible action set is nonempty and the true cost critic is available, one possible ideal construction is a projection:

\begin{align}
\label{eq4.2-1}
    \arg \min_{\Delta a}\;\;&\|\Delta a\|_1\\
s.t.\;\;&a+\Delta a \in \mathcal{A}, \;\;\Delta a \in \mathcal{A}_{sub} \notag\\
& \text{ReLU}(Q_c^{\pi}(s,a+\Delta a)-d)=0 \notag 
\end{align}
where the $\mathcal{A}_{sub}$ is the subset of the action space.

The aforementioned projection problem provides one possible way to generate a feasible reference candidate.
However, it may be infeasible, and solving it online is expensive. More importantly, the practical SGP is not intended to project the actor output into the predicted safe set in a single step. Instead, it provides a nearby reference with a lower predicted cost violation so that SGPC can guide the actor progressively toward the predicted safe set. We therefore replace the projection with a finite local search and present the sampling-based SGP below.

First, sample a series of ($n-1$) preliminary guiding actions $\Delta \hat{a}_k$ from a multivariate Gaussian with a constant diagonal covariance $\sigma_{sgp}$, i.e.,
\begin{equation}
   \Delta \hat{a}_k \sim \mathcal{N}(0, \sigma_{sgp}^2), k=1,2,...,n-1
   \label{eq4.2-2}
\end{equation}

Then clamp $\Delta \hat{a}_k$ to ensure the augmented action is in the action space's range, and we can get a new set of guiding actions.
\begin{equation}
    \Delta {a}_k = \text{clamp}(\Delta \hat{a}_k+a, \mathcal{A})-a
    \label{eq4.2-3}
\end{equation}
\begin{equation}
    A_{sgp} = [0, \Delta {a}_1, \Delta {a}_2,..., \Delta {a}_k,..., \Delta {a}_{n-1}]
    \label{eq4.2-4}
\end{equation}
Thus, the sampling-based SGP can be presented as follows: 
\begin{align}
\label{eq4.2-5}
    \pi_{sgp}^+(s,a)=\Delta a = \arg &\min_{\Delta a_k\in A_{sgp}} \left[\lambda_{sgp}\cdot\|\Delta a_k\|_1 \right.\\
    &\left.+\text{ReLU}(Q_c^{\pi}\left(s,a+\Delta a_k\right)-d)\right]\notag
\end{align}

According to Definition~\ref{def:SGP}, the sampling-based policy provides a finite-sample approximation of SGP. When $\Delta a=\mathbf{0}$, the SGPC becomes inactive.

However, in practice, there exists a gap between the estimated value ${Q}_c^{\pi}(\cdot;\theta_{c})$ (parameterized by the neural network $\theta_c$) and the true value $Q_c^{\pi}(\cdot)$, i.e.,
\begin{equation}
    {Q}_c^{\pi}(s,a;\theta_{c})={Q}_c^{\pi}(s,a)+g_c
    \label{eq4.2-6}
\end{equation}

The learned cost critic may exhibit a non-negligible aggregate estimation offset, which can shift the effective threshold used by SGP and affect whether a candidate action is classified as predicted safe, i.e., $\operatorname{ReLU}(Q_c^{\pi}(s,a+\Delta a;\theta_c)-d)=0$. Such estimation errors may arise from bootstrapped TD targets and function approximation. To reduce this aggregate offset, we use the empirical mean cumulative cost collected during the current epoch as a calibration target and estimate a batch-level centering term $\hat{g}_c$ as follows:

\begin{align}
 \label{eq4.2-7}
    \hat{g}_c = &\mathbb{E}_{s\sim\mathcal{D}_{batch},a\sim\pi}[{Q}_c^\pi(s,a;\theta)]\\&-\mathbb{E}_{(s_t,a_t)\sim\rho^{epoch}_\pi}[\sum_{t=0}^T\gamma^tc(s_t, a_t)]\notag
\end{align}

Given the assumption that $Var[g_c]\approx0$, we replace the $g_c$ in \eqref{eq4.2-6} with $\hat{g}_c$, and substitute this into \eqref{eq4.2-5}. However, the two expectations in \eqref{eq4.2-7} involve different state distributions and remaining horizons. Therefore, $\hat{g}_c$ should not be interpreted as an unbiased estimator of $g_c(s,a)$, but rather as a scalar centering heuristic used in the practical implementation.
\begin{align}
\label{eq4.2-8}
\pi_{sgp}^+(s,a)=\Delta a =& \arg \min_{\Delta a_k\in A_{sgp}} \left[\lambda_{sgp}\cdot\|\Delta a_k\|_1 \right.\\
&\left.+\text{ReLU}(Q_c^{\pi}\left(s,a+\Delta a_k;\theta\right)-\hat{g}_c-d)\right]\notag
\end{align}

Moreover, \eqref{eq4.2-8} is consequently used only during the actor update stage and does not alter the action executed in the environment.

\subsection{SGPC-based Primal-Dual Off-Policy Safe RL}
\label{sec:4.3}
This section introduces the SGPC-based off-policy safe RL algorithm based on the SGPC-based COP. 

First, we train both Q-functions for the expected cumulative reward $Q_{\theta_r}^\pi$ and cost $Q_{\theta_c}^\pi$, which are parameterized by the networks $\theta_r$ and $\theta_c$, respectively. The two critic networks are updated by minimizing the TD-errors $J_{Q_r}$ and $J_{Q_c}$, respectively, following the standard training approach in off-policy safe RL such as SAC-Lag and TD3-Lag. 

The SGPC operates solely during the actor's training process to train the policy $\pi_\phi$ parameterized by $\phi$. According to Section~\ref{sec:3.2}, for the original COP in \eqref{eq3.1-1}, Lagrangian-based off-policy algorithms update the policy $\pi_\phi$ by minimizing the actor loss $J_\pi^{origin}$ as follows:
\begin{equation}
    J_\pi^{origin}(\phi) = \mathbb{E}_{s\sim \mathcal{D}}\left[-Q_{\theta_r}^\pi(s,\pi_\phi(s))+\lambda Q_{\theta_c}^\pi(s,\pi_\phi(s)) \right]
    \label{eq4.3-1}
\end{equation}
where $\mathcal{D}$ denotes a batch of states sampled from the replay buffer, and $\lambda$ denotes the Lagrange multiplier for $Q_{\theta_c}^\pi$.

For the ideal SGPC-based COP in \eqref{eq4.1-1}, the corresponding dual problem is: 
\begin{align}
\label{eq4.3-2}
    \min_{\lambda\geq0,\mu\geq0}\max_\pi \;\; &\left\{\mathbb{E}_{s,a}\left[Q_{\theta_r}^{\pi}(s,a+\Delta a)\right]\right.\\
    &\left.-\lambda\left(\mathbb{E}_{s,a}\left[Q_{\theta_c}^{\pi}(s,a+\Delta a)\right]-d \right)\right.\notag\\
    &\left.-\mu\mathbb{E}_{s,a}\left[\|\Delta a\|_1\right] \right\}\notag
\end{align}
where $\mu$ represents the Lagrange multiplier for SGPC. According to the Lagrangian method, the loss function for the updating of the policy can be constructed as follows:
\begin{align}
    \label{eq4.3-3}
    J_\pi^{sgpc}(\phi) = \mathbb{E}_{s\sim \mathcal{D}}[&-Q_{\theta_r}^\pi(s,\pi_\phi(s)+\Delta a)\\
    &+\lambda Q_{\theta_c}^\pi(s,\pi_\phi(s)+\Delta a)\notag \\
    &+\mu \|\text{sg}(\pi_\phi(s)+\Delta a)-\pi_\phi(s)\|_1]\notag
\end{align}
where $\text{sg}(\cdot)$ is the stop gradient function. 

Furthermore, when SGP produces a nonzero guiding action, the SGPC becomes active, leading to a divergence between the original and new problems. To retain the original optimization objective while incorporating local guidance, we combine $J_\pi^{sgpc}$ and $J_\pi^{origin}$ in the final actor loss $J_\pi^{final}$ below. This mixed loss is a practical surrogate motivated by the auxiliary COP in \eqref{eq4.1-1}.

\begin{equation}
    J_\pi^{final}(\phi) = \left[\rho J_\pi^{sgpc}(\phi) + (1-\rho) J_\pi^{origin}(\phi)\right]/(1+\lambda)
    \label{eq4.3-4}
\end{equation}
where $\rho \in \left[0,1\right]$ denotes the primal-dual ratio for the policy's final loss function. The default value is set at $\rho=0.5$.

Following the update of the policy $\pi_\phi$, the Lagrangian multipliers $\lambda$ and $\mu$ are then updated iteratively by minimizing their loss functions $J_\lambda$ and $J_\mu$, respectively.
\begin{equation}
    J_\lambda(\lambda) = \lambda\mathbb{E}_{s\sim \mathcal{D}}\left[d- Q_{\theta_c}^\pi(s,\pi_\phi(s)+\Delta a) \right]
    \label{eq4.3-5}
\end{equation}
\begin{equation}
    J_\mu(\mu) = -\mu\mathbb{E}_{s\sim \mathcal{D}}\left[ \|\Delta a\|_1 \right]
    \label{eq4.3-6}
\end{equation}

\begin{algorithm}[tb]
  \caption{Generic SGPC-based Off-Policy Safe RL}
  \label{alg:example}
  \begin{algorithmic}[1]
    \STATE {\bfseries Input:} cost limit $d$, sample number $n$
    \STATE {\bfseries Initialize:} networks $\phi$, $\theta_r^\pi$, $\theta_c^\pi$; replay buffer $\mathcal{D}$; multipliers $\lambda$, $\mu$
    \FOR{each iteration}
        \FOR{each environment step}
        \STATE $a_t$ $\sim$ $\pi_{\phi}(\cdot\mid s_t)$
        \STATE $(s_{t+1},r_t,c_t)\leftarrow \mathrm{env.step}(a_t)$
        \STATE $\mathcal{D}$ $\leftarrow$ $\mathcal{D}$ $\bigcup$ $\{s_t, a_t, r_t, c_t, s_{t+1}\}$
        \ENDFOR
        \FOR{each training step}
        \STATE
        Sample a minibatch $\mathcal{B}\sim\mathcal{D}$
        \STATE $\theta_r\leftarrow\theta_r-\eta_r\nabla_{\theta_r}J_{Q_r}(\theta_r)$ // update reward Q function
        \STATE $\theta_c\leftarrow\theta_c-\eta_c\nabla_{\theta_c}J_{Q_c}(\theta_c)$ // update cost Q function
        \STATE // start update actor
        \STATE Use states $s_t$ from $\mathcal{B}$ for the actor update
        \STATE $a$ $\sim$ $\pi_{\phi}(\cdot|s_t)$
        \STATE $J_\pi^{origin}\leftarrow \mathrm{Eq.}~\eqref{eq4.3-1}(a, Q_{\theta_r}^\pi, Q_{\theta_c}^\pi,\lambda)$
        \STATE sample $A_{sgp}$ according to \eqref{eq4.2-2}-\eqref{eq4.2-4} 
        \STATE $\Delta a \leftarrow$ Eq.~\eqref{eq4.2-8}
        \STATE $J_\pi^{sgpc}\leftarrow$ Eq.~\eqref{eq4.3-3}$(a, \Delta a, Q_{\theta_r}^\pi, Q_{\theta_c}^\pi, \lambda, \mu)$
        \STATE $J_\pi^{final}\leftarrow \frac{\rho J_\pi^{sgpc}(\phi) + (1-\rho) J_\pi^{origin}(\phi)}{1+\lambda}$
        \STATE $\phi\leftarrow\phi-\eta_\pi\nabla_{\phi}J_{\pi}^{final}$
        \STATE // update Lagrange multipliers
        \STATE $\lambda\leftarrow\left[\lambda-\eta_{\lambda}\nabla_{\lambda}J_{\lambda}(\lambda)\right]_+$
        \STATE $\mu\leftarrow\left[\mu-\eta_{\mu}\nabla_{\mu}J_{\mu}(\mu)\right]_+$
        \ENDFOR
    \ENDFOR
    \label{alg:sgpc_generic}
  \end{algorithmic}
\end{algorithm}

Equation~\eqref{eq4.3-5} evaluates the SGP-guided reference, whereas environment interaction uses the original actor action. Thus, SGPC is applied only during actor updates and does not directly modify the actions executed in the environment. Moreover, when the estimated cost falls below $d$, $\lambda$ may decrease, allowing reward-driven updates to move the actor outside the predicted safe set and cause oscillations. In contrast, $\mu$ accumulates when $\Delta a\neq\mathbf{0}$ and remains unchanged when $\Delta a=\mathbf{0}$. Therefore, if a later actor update leaves the predicted safe set and SGP finds an improved guiding reference, indicating that the actor has not remained stably feasible over the sampled states,  the guidance reactivates with the retained multiplier and the SGPC immediately reactivates the guiding gradient. This is the mechanism intended to reduce repeated boundary crossings.

Overall, the generic training procedure of SGPC-based off-policy safe RL is summarized in \textbf{Algorithm 1}. SGPC follows the standard actor-critic training loop while modifying only the actor update. The SGPC-specific steps are concentrated in Lines~17-24. This block can be readily integrated into different off-policy primal-dual backbones.

\subsection{Two Instantiations: SAC-SGPC and TD3-SGPC}
\label{sec:4.4}
We instantiate SGPC in two representative off-policy safe RL algorithms, SAC-Lag and TD3-Lag, yielding SAC-SGPC and TD3-SGPC, respectively. Below, we describe how SGPC is incorporated into the algorithms.

\subsubsection{SAC-SGPC}
Compared with SAC-Lag, SAC-SGPC modifies only the actor objective and associated multiplier updates while retaining the critic's training procedure. As in SAC-Lag, it employs two reward Q-functions, $Q^\pi_{\theta_{r1}}$ and $Q^\pi_{\theta_{r2}}$, and one cost Q-function $Q^\pi_{\theta_c}$. The original actor loss $J_\pi^{origin}$ can be seen as follows:

\begin{align}
J_\pi^{\mathrm{origin}}(\phi)
= \mathbb{E}_{s\sim\mathcal{D}}\Big[
& \alpha\log \pi_\phi(a_\phi|s)
- \min_{i=1,2}\{Q_{\theta_{r_i}}^\pi(s,a_\phi)\} \notag \\
& + \lambda Q_{\theta_c}^\pi(s,a_\phi)
\Big],
\label{eqa.2-1}
\end{align}
where $a_\phi$ is sampled from the policy distribution parameterized by $\phi$, and $\alpha$ is the entropy temperature inherited from the SAC-Lag backbone.

The implementation retains the entropy term evaluated at the actor-sampled action $a_\phi$, while evaluating the reward and cost critics at the SGP-guided reference action $a_\phi+\Delta a$. All remaining steps follow those described in Section~\ref{sec:4.3}, and the SGPC actor loss $J_\pi^{sgpc}$ is defined as follows:

\begin{align}
J_\pi^{\mathrm{sgpc}}(\phi)
= \mathbb{E}_{s\sim\mathcal{D}}\Big[
& \alpha\log \pi_\phi(a_\phi|s)
- \min_{i=1,2}\{Q_{\theta_{r_i}}^\pi(s,a_\phi+\Delta a)\} \notag \\
& + \lambda Q_{\theta_c}^\pi(s,a_\phi+\Delta a) \notag \\
& + \mu \|\mathrm{sg}(a_\phi+\Delta a)-a_\phi\|_1
\Big].
\label{eqa.2-2}
\end{align}

\subsubsection{TD3-SGPC}
Although TD3 employs two reward Q-functions, only one reward Q-function is used during the actor update. Therefore, the SGPC algorithm can be directly incorporated into the actor update process following the steps in Section~\ref{sec:4.3} to implement TD3-SGPC.

\section{Experiments}
\label{sec:5}
We evaluate SAC-SGPC and TD3-SGPC on multiple tasks from the Safety-Gymnasium benchmark~\cite{Safety_Gymnasium_2023}. Section~\ref{sec:5.1} describes the experimental setup. Section~\ref{sec:5.2} compares the proposed methods with five representative off-policy safe RL baselines under the default cost limit, while Section~\ref{sec:5.3} evaluates their performance under tighter cost limits. Section~\ref{sec:5.4} presents comparisons with on-policy safe RL methods. Finally, Sections~\ref{sec:5.5} and~\ref{sec:5.6} report ablation studies and hyperparameter analyses, respectively.
\subsection{Experiment Settings}
\label{sec:5.1}
\begin{figure*}[!tb]
  \begin{center}
    \centerline{\includegraphics[width=2\columnwidth]{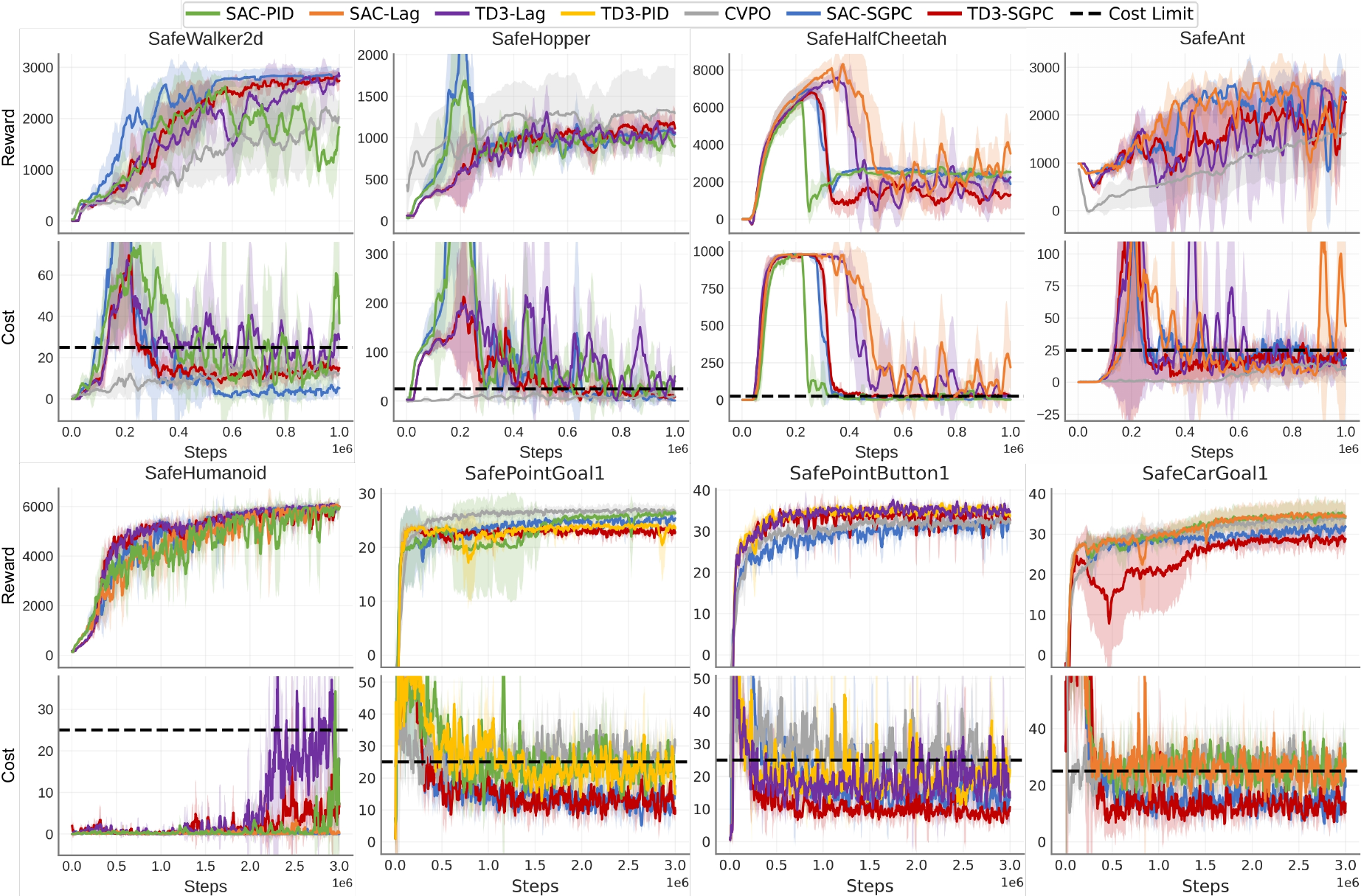}}
    \caption{Reward and cost learning curves of comparative experiments under the default cost limit of 25. For readability, each task shows the two SGPC-based methods and the three best-performing baselines for that task.}
    \label{exp-cost25}
  \end{center}
\end{figure*}

We conduct experiments on eight Safety-Gymnasium tasks~\cite{Safety_Gymnasium_2023}, including velocity-constrained locomotion and safe navigation tasks. For brevity, we use shortened task names throughout the paper. The correspondence between full Safety-Gymnasium environment names and shortened names is summarized in Table~\ref{tab:env_names}. We use five random seeds, $\{0,1,2,3,4\}$. The maximum episode length is 1000. The total number of environment steps is set to $1\times10^6$ for the first four tasks listed in Table~\ref{tab:env_names} and $3\times10^6$ for the last four tasks.

\begin{table}[tb]
\caption{Summary of environments used in our experiments}
\centering
\begin{tabular}{ll}
\toprule
Environment & Full Environment Name \\
\midrule
SafeAnt  & SafetyAntVelocity-v1 \\
SafeHalfCheetah  & SafetyHalfCheetahVelocity-v1 \\
SafeHopper  & SafetyHopperVelocity-v1 \\
SafeWalker2d  & SafetyWalker2dVelocity-v1 \\
SafeHumanoid  & SafetyHumanoidVelocity-v1 \\
SafePointGoal1  & SafetyPointGoal1-v0 \\
SafeCarGoal1  & SafetyCarGoal1-v0 \\
SafePointButton1  & SafetyPointButton1-v0 \\
\bottomrule
\end{tabular}
\label{tab:env_names}
\end{table}

The default cost limit is set to 25 for all environments, and tighter cost limits of 15 and 5 are further considered to evaluate performance under more stringent safety constraints. Each method is evaluated with five random seeds, and all reported curves are averaged over the five seeds, with shaded regions denoting the standard deviation. To evaluate the trained policies, we run 500 evaluation episodes for each method after training, consisting of 100 episodes for each of the five trained seeds. For each trained policy, evaluation episodes are generated with non-fixed environment seeds. Evaluation results in tables are reported as \textbf{mean $\pm$ standard deviation}. For each metric in Table~\ref{table_sac_25} and Table~\ref{table_sac_tighter}, we separately \textbf{bold the top two mean values and the two smallest standard deviations}, where higher rewards, lower costs, and smaller standard deviations are preferred.

We compare SAC-SGPC and TD3-SGPC with five representative off-policy safe RL baselines: SAC-Lag and TD3-Lag, two Lagrangian-based algorithms~\cite{Lag_RL-Ha_2021,Lag_RL_Ray_2019}; SAC-PID and TD3-PID, two PID Lagrangian-based algorithms~\cite{PIDLag_2020}; and CVPO~\cite{CVPO}. For a fair comparison, all baselines are evaluated under the same environment settings. All policies are trained on NVIDIA A800 GPUs. Detailed information about the environments, implementation details, and hyperparameter settings is provided in \textbf{Appendix~\ref{App:C}.}

\begin{table*}[!tb]
\caption{Evaluation results on Safety-Gymnasium tasks under the default cost limit of 25}
\label{table_sac_25}
\setlength{\tabcolsep}{2pt}
\centering
\begin{tabular}{@{}ll*{8}{c}@{}}
\toprule
\multirow{2}{*}{\textbf{Algorithms}}
& \multirow{2}{*}{\textbf{Metrics}}
& \multicolumn{8}{c}{\textbf{Environment}} \\
\cmidrule(lr){3-10}
& & SafeAnt & SafeHalfCheetah & SafeHopper & SafeWalker2d
  & SafeHumanoid & SafePointGoal1 & SafeCarGoal1 & SafePointButton1 \\
\midrule
\multirow{2}{*}{SAC-Lag}
& Reward
& 1331.2$\pm$1788.7
& \textbf{3321.7}$\pm$3430.0
& 931.7$\pm$156.2
& 920.5$\pm$535.6
& 5974.3$\pm$362.4
& 20.6$\pm$10.3
& \textbf{34.8}$\pm$4.4
& 32.6$\pm$4.8 \\
& Cost
& 22.1$\pm$32.9
& 195.3$\pm$389.3
& 17.1$\pm$34.1
& 27.1$\pm$36.9
& \textbf{0.5}$\pm$4.3
& 21.3$\pm$24.4
& 25.9$\pm$25.5
& 26.9$\pm$46.7 \\
\cmidrule(lr){2-10}
\multirow{2}{*}{TD3-Lag}
& Reward
& \textbf{2539.9}$\pm$\textbf{362.9}
& 2186.4$\pm$170.1
& 1095.7$\pm$344.1
& 2679.9$\pm$691.3
& \textbf{6085.4}$\pm$497.2
& 23.5$\pm$3.0
& 29.9$\pm$\textbf{2.8}
& \textbf{34.7}$\pm$5.0 \\
& Cost
& 20.0$\pm$16.3
& 9.9$\pm$\textbf{5.4}
& 72.9$\pm$111.1
& 14.1$\pm$19.6
& 28.6$\pm$40.1
& 24.1$\pm$22.0
& 27.3$\pm$25.3
& 16.5$\pm$29.4 \\
\cmidrule(lr){2-10}
\multirow{2}{*}{SAC-PID}
& Reward
& 1617.6$\pm$1557.8
& 2520.7$\pm$390.1
& \textbf{1128.1}$\pm$\textbf{230.3}
& 1608.2$\pm$981.3
& 5697.1$\pm$965.7
& \textbf{26.2}$\pm$\textbf{1.7}
& \textbf{34.8}$\pm$4.0
& 33.2$\pm$4.6 \\
& Cost
& 27.9$\pm$42.3
& \textbf{3.0}$\pm$6.7
& 34.1$\pm$68.1
& 17.1$\pm$22.1
& 8.9$\pm$25.2
& 26.5$\pm$24.0
& 23.1$\pm$24.6
& 22.0$\pm$27.4 \\
\cmidrule(lr){2-10}
\multirow{2}{*}{TD3-PID}
& Reward
& 1506.2$\pm$1581.6
& 2162.6$\pm$313.6
& 1058.0$\pm$312.6
& 1791.7$\pm$1059.1
& 5926.4$\pm$1082.3
& 23.9$\pm$1.9
& 30.7$\pm$3.4
& \textbf{35.2}$\pm$\textbf{4.5} \\
& Cost
& 13.8$\pm$20.6
& 19.6$\pm$15.3
& 32.4$\pm$48.2
& 15.0$\pm$21.6
& 23.6$\pm$23.3
& 23.1$\pm$21.8
& 25.3$\pm$25.7
& 20.5$\pm$25.7 \\
\cmidrule(lr){2-10}
\multirow{2}{*}{CVPO}
& Reward
& 1659.2$\pm$744.5
& 1988.1$\pm$458.4
& \textbf{1244.3}$\pm$549.7
& 2144.4$\pm$879.4
& 3006.4$\pm$1606.9
& \textbf{26.8}$\pm$1.8
& 34.3$\pm$\textbf{2.5}
& 32.4$\pm$\textbf{4.4} \\
& Cost
& 10.7$\pm$15.3
& 19.2$\pm$12.7
& 10.8$\pm$10.2
& 14.8$\pm$15.0
& 2.1$\pm$5.8
& 24.9$\pm$25.3
& 23.5$\pm$25.6
& 25.1$\pm$36.6 \\
\cmidrule(lr){2-10}
\multirow{2}{*}{\textbf{SAC-SGPC}}
& Reward
& \textbf{2886.0}$\pm$\textbf{96.4}
& \textbf{2658.0}$\pm$\textbf{161.6}
& 1120.8$\pm$\textbf{84.9}
& \textbf{2831.3}$\pm$\textbf{111.1}
& \textbf{6005.9}$\pm$\textbf{268.6}
& 25.6$\pm$\textbf{1.4}
& 31.8$\pm$4.5
& 29.4$\pm$9.2 \\
& Cost
& \textbf{7.0}$\pm$\textbf{8.7}
& \textbf{0.0}$\pm$\textbf{0.0}
& \textbf{4.0}$\pm$\textbf{6.9}
& \textbf{1.5}$\pm$\textbf{3.1}
& \textbf{0.0}$\pm$\textbf{0.0}
& \textbf{12.6}$\pm$\textbf{16.7}
& \textbf{13.0}$\pm$\textbf{18.5}
& \textbf{14.4}$\pm$\textbf{19.6} \\
\cmidrule(lr){2-10}
\multirow{2}{*}{\textbf{TD3-SGPC}}
& Reward
& 2201.1$\pm$455.7
& 1997.6$\pm$\textbf{140.2}
& 826.9$\pm$521.1
& \textbf{2850.7}$\pm$\textbf{175.4}
& 5926.4$\pm$\textbf{271.1}
& 23.3$\pm$2.4
& 29.5$\pm$4.1
& 33.8$\pm$5.1 \\
& Cost
& \textbf{5.0}$\pm$\textbf{7.9}
& 15.6$\pm$11.2
& \textbf{0.7}$\pm$\textbf{1.3}
& \textbf{11.7}$\pm$\textbf{7.2}
& 0.8$\pm$\textbf{2.1}
& \textbf{9.6}$\pm$\textbf{12.6}
& \textbf{14.4}$\pm$\textbf{18.2}
& \textbf{8.8}$\pm$\textbf{8.5} \\
\bottomrule
\end{tabular}
\end{table*}

\begin{table*}[!tb]
\caption{Evaluation results on selected representative tasks under tighter cost limits of 5 and 15}
\centering
\begin{tabular}{cclllll}
\toprule
\multirow{2}{*}{\textbf{Cost Limit}} 
& \multirow{2}{*}{\textbf{Algorithms}} 
& \multirow{2}{*}{\textbf{Metrics}} 
& \multicolumn{4}{c}{\textbf{Environment}} \\
\cmidrule(lr){4-7}
\multicolumn{3}{c}{} & SafeAnt & SafeHopper & SafeWalker2d & SafePointGoal1 \\
\midrule
\multirow{14}{*}{5}
& SAC-Lag & Reward & 1953.12$\pm$1148.35 & 755.5$\pm$510.67 & 577.97$\pm$474.69 & 17.6$\pm$10.0 \\
&        & Cost   & \textbf{0.45}$\pm$1.96       & 32.69$\pm$65.06  & 10.39$\pm$26.8    & 5.1$\pm$10.4 \\
\cmidrule(lr){3-7}
& TD3-Lag & Reward & 1266.58$\pm$1233.02 & 966.32$\pm$77.84 & 2270.41$\pm$643.58 & 22.7$\pm$2.7 \\
&        & Cost   & 10.9$\pm$21.69      & 390.2$\pm$117.55 & 7.522$\pm$14.92    & 4.1$\pm$7.6 \\
\cmidrule(lr){3-7}
& SAC-PID & Reward & 1263.03$\pm$1431.68 & 824.15$\pm$216.44 & 838.99$\pm$660.7 & 11.4$\pm$13.7 \\
&        & Cost   & 6.82$\pm$16.75      & 58.14$\pm$86.71   & 12.8$\pm$17.9    & \textbf{2.9}$\pm$8.6 \\
\cmidrule(lr){3-7}
& TD3-PID & Reward & \textbf{2526.72}$\pm$\textbf{275.05} & 925.81$\pm$460.72 & 1624.8$\pm$811.82 & \textbf{22.8}$\pm$2.8 \\
&        & Cost   & 7.42$\pm$9.32      & 79.35$\pm$78.32   & 14.18$\pm$18.89   & 3.4$\pm$7.0 \\
\cmidrule(lr){3-7}
& CVPO   & Reward & 1971.7$\pm$553.9 & 698.0$\pm$488.0 & 1286.5$\pm$837.9 & \textbf{25.8}$\pm$\textbf{2.2} \\
&        & Cost   & 4.2$\pm$3.7      & 2.0$\pm$3.6     & 3.4$\pm$6.2      & 7.3$\pm$11.7 \\
\cmidrule(lr){3-7}
& \textbf{SAC-SGPC} & Reward & \textbf{2828.9}$\pm$\textbf{56.23} & \textbf{1045.89}$\pm$\textbf{2.37} & \textbf{2601.43}$\pm$\textbf{265.54} & 18.7$\pm$9.7 \\
&                   & Cost   & \textbf{0.59}$\pm$\textbf{1.2}     & \textbf{0}$\pm$\textbf{0}           & \textbf{0.08}$\pm$\textbf{1.55}      & \textbf{1.8}$\pm$\textbf{5.3} \\
\cmidrule(lr){3-7}
& \textbf{TD3-SGPC} & Reward & 2512.93$\pm$440.43 & \textbf{1059.36}$\pm$\textbf{44.08} & \textbf{2717.71}$\pm$\textbf{197.87} & 22.1$\pm$\textbf{2.3} \\
&                   & Cost   & 0.95$\pm$\textbf{1.52}      & \textbf{0.47}$\pm$\textbf{2.5}      & \textbf{2.28}$\pm$\textbf{2.71}      & 3.0$\pm$\textbf{6.6} \\
\midrule
\multirow{14}{*}{15}
& SAC-Lag & Reward & 963.6$\pm$1358.32 & 672.94$\pm$429.97 & 1229.62$\pm$718.95 & 17.5$\pm$10.0 \\
&        & Cost   & \textbf{0.972}$\pm$\textbf{1.85}    & 22.55$\pm$43.74   & 23.37$\pm$34.61    & 11.7$\pm$16.8 \\
\cmidrule(lr){3-7}
& TD3-Lag & Reward & 1525.46$\pm$1365.27 & 1123.89$\pm$185.39 & 1871.38$\pm$801.24 & 23.9$\pm$\textbf{1.6} \\
&        & Cost   & 43.82$\pm$73.47     & 7.27$\pm$27.17     & 17.59$\pm$25.27    & 15.8$\pm$17.8 \\
\cmidrule(lr){3-7}
& SAC-PID & Reward & \textbf{2829.25}$\pm$\textbf{206.1} & 450.97$\pm$325.56 & 968.12$\pm$822.09 & 15.0$\pm$12.7 \\
&        & Cost   & 5.41$\pm$6.27     & 32.55$\pm$40.6    & 12.85$\pm$21.13   & 8.7$\pm$15.9 \\
\cmidrule(lr){3-7}
& TD3-PID & Reward & 1967.04$\pm$758.42 & 1069.77$\pm$162.62 & 2644.18$\pm$365.9 & \textbf{24.4}$\pm$\textbf{1.6} \\
&        & Cost   & 28.19$\pm$47.6     & 28.06$\pm$62.49    & 19.21$\pm$31.64   & 11.5$\pm$14.3 \\
\cmidrule(lr){3-7}
& CVPO   & Reward & 1721.0$\pm$464.0 & \textbf{1355.2}$\pm$495.5 & 1461.0$\pm$612.9 & \textbf{26.4}$\pm$1.6 \\
&        & Cost   & 24.6$\pm$50.1    & 4.6$\pm$9.5      & 6.7$\pm$7.5      & 17.8$\pm$18.9 \\
\cmidrule(lr){3-7}
& \textbf{SAC-SGPC} & Reward & \textbf{2759.75}$\pm$\textbf{59.81} & \textbf{1137.67}$\pm$\textbf{86.21} & \textbf{2790.11}$\pm$\textbf{149.25} & 19.4$\pm$10.0 \\
&                   & Cost   & \textbf{0.11}$\pm$\textbf{0.34}     & \textbf{3.25}$\pm$\textbf{5.38}     & \textbf{1.84}$\pm$\textbf{3.1}       & \textbf{4.1}$\pm$\textbf{8.7} \\
\cmidrule(lr){3-7}
& \textbf{TD3-SGPC} & Reward & 2157.8$\pm$632.89 & 1070.58$\pm$\textbf{69.45} & \textbf{2764.07}$\pm$\textbf{127.41} & 22.7$\pm$1.9 \\
&                   & Cost   & 11.77$\pm$13.56   & \textbf{0.31}$\pm$\textbf{0.78}     & \textbf{5.89}$\pm$\textbf{5.97}      & \textbf{6.3}$\pm$\textbf{9.6} \\
\bottomrule
\end{tabular}
\label{table_sac_tighter}
\end{table*}

\subsection{Comparative Experiments}
\label{sec:5.2}
The learning curves under the default cost limit of 25 are shown in Fig.~\ref{exp-cost25}. According to the test cost curves, SAC-SGPC and TD3-SGPC generally maintain their costs below the predefined cost limit after convergence, demonstrating stronger empirical constraint satisfaction than the baseline methods. In addition, the shaded regions of SGPC-based methods are generally smaller, indicating lower variability across random seeds. By incorporating SGPC, the proposed methods substantially reduce the cost oscillations commonly observed in Lagrangian-based methods near the constraint boundary. Upon reaching the cost limit line, SGPC-based methods tend to remain below it, indicating improved empirical training stability.

SGPC-based methods are also among the earliest methods to reduce the test costs below the limit in most environments. This observation is consistent with the intended mechanism of guiding the actor toward the predicted safe set during actor updates. For example, in SafeWalker2d, after the cost stabilizes below the limit, the reward curves of SGPC-based methods rapidly increase to high values and remain stable, outperforming the baselines. These results indicate that SGPC can improve empirical safety while maintaining strong task performance.

The final evaluation results are reported in Table~\ref{table_sac_25}. Overall, SGPC-based methods achieve substantially lower costs and smaller cost standard deviations across most environments while maintaining competitive rewards. In particular, SAC-SGPC obtains strong reward performance on SafeAnt, SafeHalfCheetah, SafeWalker2d, and SafeHumanoid, with costs consistently below the prescribed limit. TD3-SGPC also improves constraint satisfaction compared with TD3-Lag and TD3-PID, especially on SafeHopper and the navigation tasks. 

These results indicate that SGPC improves the empirical reward-constraint trade-off across both velocity-constrained locomotion and navigation tasks. Compared with Lagrangian and PID-Lagrangian baselines, SGPC-based methods generally reduce cost violations and cost variability, suggesting more stable constraint satisfaction after training.

\subsection{Experiments under Tighter Safety Constraints}
\label{sec:5.3}
To further examine performance under tighter safety constraints, we conduct additional experiments under cost limits of 5 and 15 on four selected representative tasks. Results can be seen in Table~\ref{table_sac_tighter}. As the cost limit decreases, the advantage of SGPC-based methods becomes more pronounced in several tasks: Lagrangian-based baselines tend to exhibit larger oscillations around the constraint boundary, whereas SGPC-based methods more consistently maintain their costs near or below the limit. 

\begin{figure}[tb]
  \begin{center}
    \centerline{\includegraphics[width=1\columnwidth]{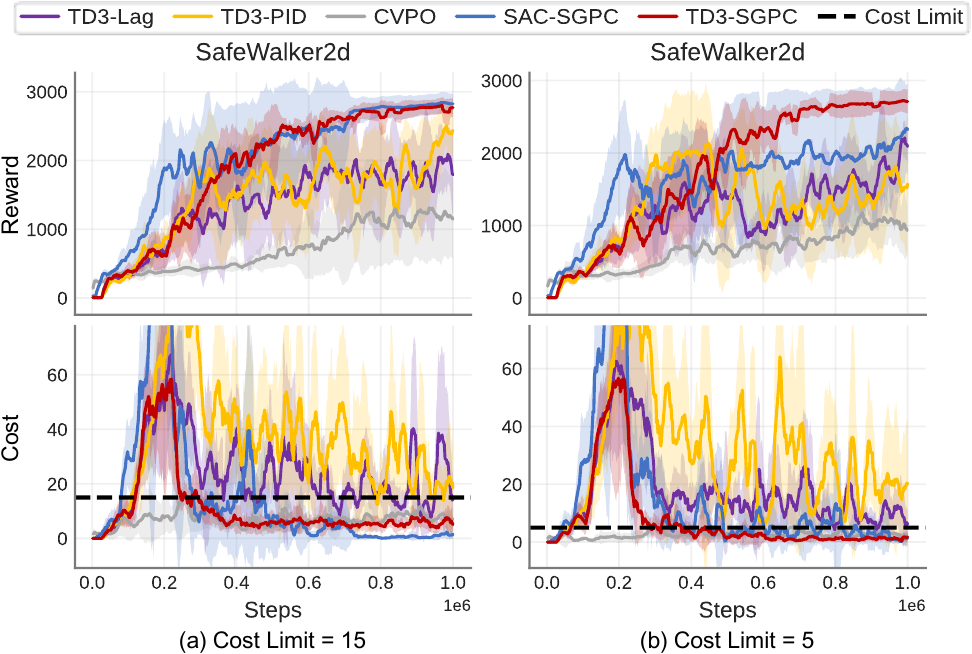}}
    \caption{Reward and cost learning curves on SafeWalker2d under tighter cost limits. }
    \label{fig:exp-cost-walk}
  \end{center}
\end{figure}

\begin{figure}[tb]
  \begin{center}
    \centerline{\includegraphics[width=1\columnwidth]{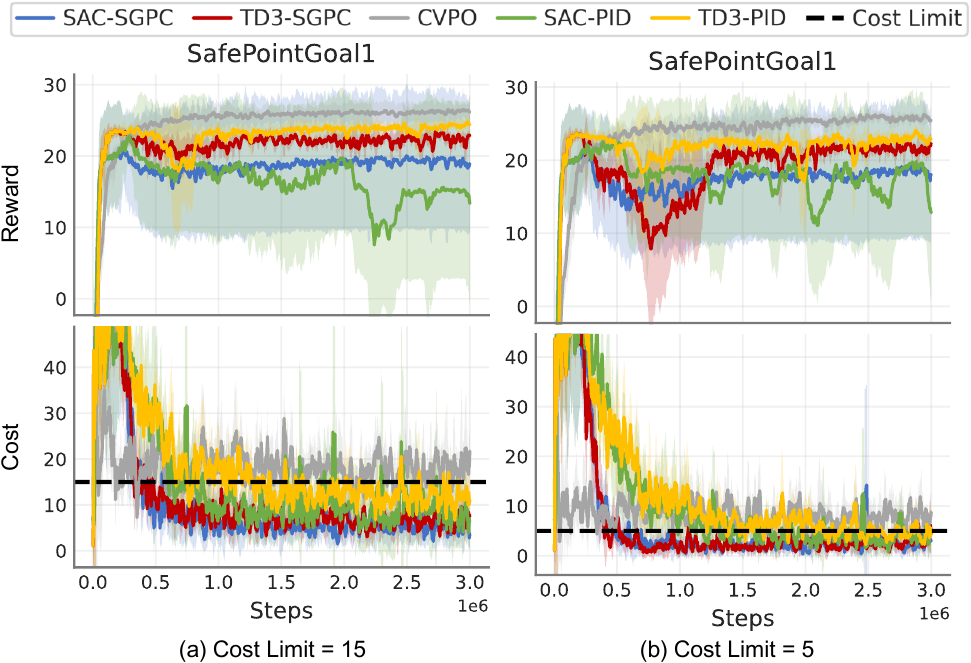}}
    \caption{Reward and cost learning curves on SafePointGoal1 under tighter cost limits.}
    \label{fig:exp-cost-point}
  \end{center}
\end{figure}

Fig.~\ref{fig:exp-cost-walk} and Fig.~\ref{fig:exp-cost-point} further illustrate the training curves on SafeWalker2d and SafePointGoal1, representing locomotion and navigation tasks, respectively. The curves show that SGPC-based methods can reduce cost violations and generally stabilize near or below the prescribed cost limits while maintaining competitive rewards, especially when the safety constraint becomes more stringent.

\subsection{Comparisons with On-Policy Safe RL Methods}
\label{sec:5.4}
In this paper, we additionally compare SAC-SGPC and TD3-SGPC with two on-policy safe RL methods, CPO and CUP. This comparison uses the same number of environment interactions. SGPC is designed for off-policy primal–dual safe RL. The on-policy methods may require a larger interaction budget to reach stable performance.


\begin{figure*}[!tb]
  \begin{center}
    \centerline{\includegraphics[width=2\columnwidth]{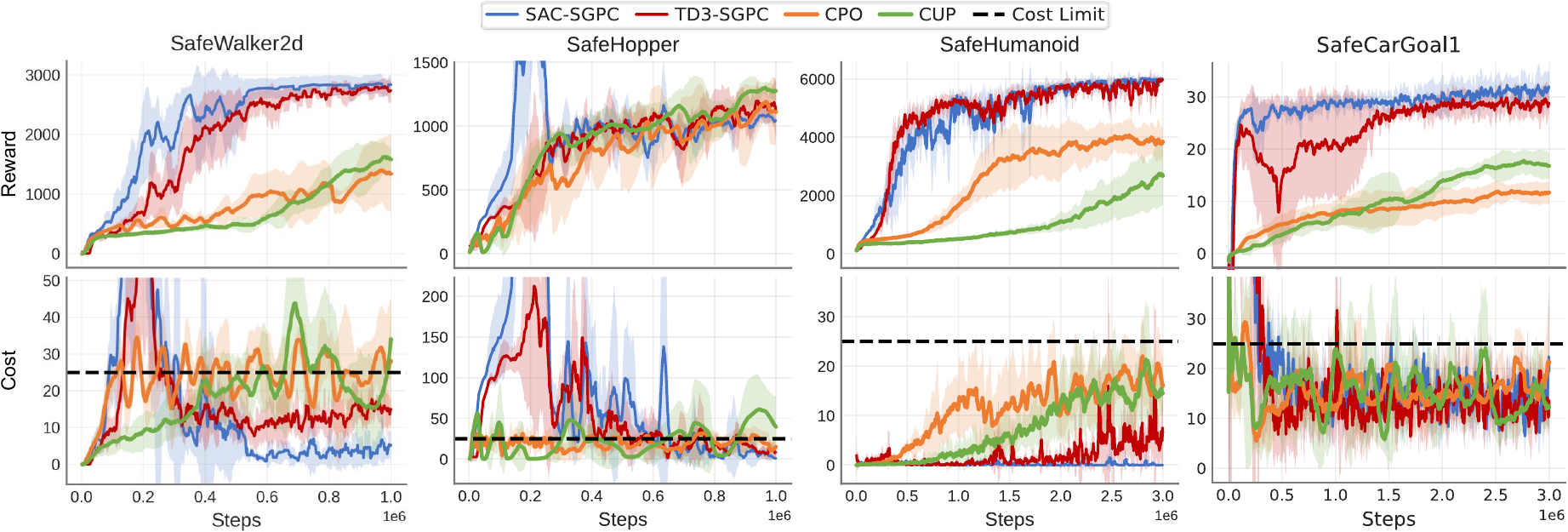}}
    \caption{Reward and cost learning curves comparing SGPC-based off-policy methods with representative on-policy safe RL methods}
    \label{fig:on-policy}
  \end{center}
\end{figure*}

As shown in Table~\ref{table_on_policy} and Fig.~\ref{fig:on-policy}, the learning curves of CPO and CUP continue to improve on several tasks under the same training budget, indicating that they may require more environment steps to fully stabilize. In contrast, SGPC-based methods achieve high task returns and low costs within the same number of environment interactions. 
\begin{table*}[!tb]
\caption{Comparison with representative on-policy safe RL methods under the default cost limit of 25.}
\label{table_on_policy}
\setlength{\tabcolsep}{2pt}
\centering
\begin{tabular}{@{}ll*{8}{c}@{}}
\toprule
\multirow{2}{*}{\textbf{Algorithms}}
& \multirow{2}{*}{\textbf{Metrics}}
& \multicolumn{8}{c}{\textbf{Environment}} \\
\cmidrule(lr){3-10}
& & SafeAnt & SafeHalfCheetah & SafeHopper & SafeWalker2d
  & SafeHumanoid & SafePointGoal1 & SafeCarGoal1 & SafePointButton1 \\
\midrule
\multirow{2}{*}{CPO}
& Reward
& 2368.8$\pm$586.7
& 1802.2$\pm$202.8
& 1105.0$\pm$428.5
& 1520.3$\pm$665.8
& 5393.4$\pm$1282.5
& 7.1$\pm$5.8
& 12.1$\pm$6.1
& 12.0$\pm$5.2 \\
& Cost
& 35.2$\pm$44.0
& 30.4$\pm$25.7
& 61.1$\pm$87.1
& 44.9$\pm$87.9
& 1.74$\pm$6.28
& 19.4$\pm$41.4
& 29.6$\pm$108.5
& 31.1$\pm$36.4 \\
\cmidrule(lr){2-10}
\multirow{2}{*}{CUP}
& Reward
& 2164.0$\pm$406.8
& 1727.4$\pm$428.8
& \textbf{1385.9}$\pm$233.6
& 2135.7$\pm$597.1
& 3965.8$\pm$1905.2
& 9.9$\pm$5.2
& 17.2$\pm$5.8
& 6.9$\pm$\textbf{4.8} \\
& Cost
& 51.5$\pm$70.1
& 17.8$\pm$19.1
& 10.9$\pm$14.7
& 119.4$\pm$119.2
& 13.79$\pm$28.1
& 18.5$\pm$24.6
& \textbf{11.9}$\pm$23.9
& 24.4$\pm$59.3 \\
\cmidrule(lr){2-10}
\multirow{2}{*}{\textbf{SAC-SGPC}}
& Reward
& \textbf{2886.0}$\pm$\textbf{96.4}
& \textbf{2658.0}$\pm$161.6
& 1120.8$\pm$\textbf{84.9}
& 2831.3$\pm$\textbf{111.1}
& \textbf{6005.9}$\pm$\textbf{268.6}
& \textbf{25.6}$\pm$\textbf{1.4}
& \textbf{31.8}$\pm$4.5
& 29.4$\pm$9.2 \\
& Cost
& 7.0$\pm$8.7
& \textbf{0.0}$\pm$\textbf{0.0}
& 4.0$\pm$6.9
& \textbf{1.5}$\pm$\textbf{3.1}
& \textbf{0.0}$\pm$\textbf{0.0}
& 12.6$\pm$16.7
& 13.0$\pm$18.5
& 14.4$\pm$19.6 \\
\cmidrule(lr){2-10}
\multirow{2}{*}{\textbf{TD3-SGPC}}
& Reward
& 2201.1$\pm$455.7
& 1997.6$\pm$\textbf{140.2}
& 826.9$\pm$521.1
& \textbf{2850.7}$\pm$175.4
& 5926.4$\pm$271.1
& 23.3$\pm$2.4
& 29.5$\pm$\textbf{4.1}
& \textbf{33.8}$\pm$5.1 \\
& Cost
& \textbf{5.0}$\pm$\textbf{7.9}
& 15.6$\pm$11.2
& \textbf{0.7}$\pm$\textbf{1.3}
& 11.7$\pm$7.2
& 0.8$\pm$2.1
& \textbf{9.6}$\pm$\textbf{12.6}
& 14.4$\pm$\textbf{18.2}
& \textbf{8.8}$\pm$\textbf{8.5} \\
\bottomrule
\end{tabular}
\end{table*}

\subsection{Ablation Studies}
\label{sec:5.5}
To investigate the contribution of each component in SGPC-based methods, we conduct ablation studies on three representative environments and compare the full SGPC-based methods with three variants:

\begin{itemize}
    \item SGPC w/o $J_\pi^{\mathrm{sgpc}}(\phi)$: removing the SGPC policy objective $J_\pi^{\mathrm{sgpc}}(\phi)$ by setting $\rho=0$.
    \item SGPC w/o $J_\pi^{\mathrm{origin}}(\phi)$: removing the original policy objective $J_\pi^{\mathrm{origin}}(\phi)$ by setting $\rho=1$.
    \item SGPC w/o $\|\Delta a\|_1$: retaining only the SGP component while removing the SGPC regularizer by deleting the term $\mu \|\mathrm{sg}(\pi_\phi(s)+\Delta a)-\pi_\phi(s)\|_1$ from $J_\pi^{\mathrm{sgpc}}(\phi)$.
\end{itemize}

The results are reported in Table~\ref{table_ablation_algo}. Removing $J_\pi^{\mathrm{sgpc}}(\phi)$ generally leads to higher mean cost or larger cost standard deviations, supporting the importance of both SGP and SGPC. Removing $J_\pi^{\mathrm{origin}}(\phi)$ often reduces reward performance or destabilizes constraint satisfaction. Removing the $\|\Delta a\|_1$ regularizer also degrades performance across most reported reward and cost metrics, particularly on SafeAnt and SafeWalker2d. In contrast, the full SGPC-based methods generally achieve low costs while maintaining competitive rewards, indicating that these components play complementary roles in safe policy optimization.

\begin{table*}[!tb]
\caption{Ablation results of SGPC components under the default cost limit of 25}
\centering
\begin{tabular}{clcccccc}
\toprule
\multirow{2}{*}{\textbf{Variants}} 
& \multirow{2}{*}{\textbf{Metrics}} 
& \multicolumn{3}{c}{\textbf{SAC-SGPC}} 
& \multicolumn{3}{c}{\textbf{TD3-SGPC}} \\
\cmidrule(lr){3-5} \cmidrule(lr){6-8}
\multicolumn{2}{c}{} 
& SafeAnt & SafeWalker2d & SafePointGoal1 
& SafeAnt & SafeWalker2d & SafePointGoal1 \\
\midrule
\multirow{2}{*}{w/o $J_\pi^{\mathrm{sgpc}}(\phi)$}
& Reward & 1331.2$\pm$1788.7 & 917.4$\pm$509.1  & 20.5$\pm$10.3
         & \textbf{2543.5}$\pm$\textbf{357.1}  & 2654.8$\pm$729.7 & \textbf{23.5}$\pm$3.0 \\
& Cost   & 22.1$\pm$32.9     & 26.4$\pm$37.3    & 22.4$\pm$26.4
         & 20.3$\pm$16.4     & 12.4$\pm$16.3    & 24.1$\pm$22.0 \\
\midrule
\multirow{2}{*}{w/o $J_\pi^{\mathrm{origin}}(\phi)$}
& Reward & 2261.6$\pm$1057.2 & \textbf{2869.1}$\pm$192.2 & 20.2$\pm$10.2
         & 1814.5$\pm$1104.9 & 2749.2$\pm$313.8 & 23.2$\pm$\textbf{1.7} \\
& Cost   & \textbf{1.5}$\pm$\textbf{2.0}       & 51.3$\pm$101.8   & 13.0$\pm$18.8
         & 50.6$\pm$89.9     & 17.2$\pm$14.1    & 11.7$\pm$13.9 \\
\midrule
\multirow{2}{*}{w/o $\|\Delta a\|_1$}
& Reward & 1580.0$\pm$1213.2 & 860.7$\pm$377.9  & 20.5$\pm$10.3
         & 902.7$\pm$1807.1  & 2814.7$\pm$650.6 & 23.3$\pm$2.3 \\
& Cost   & 15.5$\pm$30.0     & 14.0$\pm$33.8    & 20.0$\pm$24.2
         & 30.2$\pm$46.4     & 27.6$\pm$16.0    & 26.1$\pm$24.5 \\
\midrule
\multirow{2}{*}{\textbf{Full SGPC}}
& Reward & {\textbf{2886.0}}$\pm${\textbf{96.4}} & 2831.3$\pm$\textbf{111.1} & \textbf{25.6$\pm$1.4}
         & 2201.1$\pm$455.7        & {\textbf{2850.7}$\pm$\textbf{175.4}} & 23.3$\pm$2.4 \\
& Cost   & {7.0$\pm$8.7}     & {\textbf{1.5}$\pm$\textbf{3.1}}     & {\textbf{12.6}$\pm$\textbf{16.7}}
         & {\textbf{5.0}$\pm$\textbf{7.9}}     & {\textbf{11.7}$\pm$\textbf{7.2}}     & {\textbf{9.6}$\pm$\textbf{12.6}} \\
\bottomrule
\end{tabular}
\label{table_ablation_algo}
\end{table*}

\subsection{Hyperparameter Analyses}
\label{sec:5.6}
We further analyze the sensitivity of SGPC to two key hyperparameters: the SGP sampling range $\sigma_{\mathrm{sgp}}$ and the sample number $n$. The sampling range controls the scale of local action perturbations used by SGP, while the sample number determines how many candidate perturbations are evaluated during each actor update.

Table~\ref{table_ablation_range} reports the results under different values of $\sigma_{\mathrm{sgp}}$. Overall, SGPC is relatively insensitive to the sampling range. Across $\sigma_{\mathrm{sgp}}\in\{0.02,0.1,0.5\}$, SGPC-based methods generally maintain low costs and competitive rewards on the evaluated tasks, suggesting that the proposed method does not rely on a finely tuned sampling range. We use $\sigma_{\mathrm{sgp}}=0.1$ as the default setting because it provides consistently good performance across both SAC-SGPC and TD3-SGPC without task-specific tuning.

Table~\ref{table_ablation_num} reports the results under different sample numbers $n$. Overall, SGPC does not exhibit a monotonic improvement as $n$ increases. Although a larger $n$ provides more candidate perturbations for constructing the SGP reference, it also introduces more cost-critic evaluations during each actor update. Across the results, a small sample number already achieves competitive rewards and low costs on most tasks, suggesting that SGPC does not require an exhaustive search over action perturbations. Therefore, we use $n=16$ for locomotion tasks and $n=64$ for navigation tasks as the default settings.

\begin{table*}[!tb]
\caption{Hyperparameter analysis of the SGP sampling range $\sigma_{\mathrm{sgp}}$}
\centering
\begin{tabular}{clcccccc}
\toprule
\multirow{2}{*}{$\boldsymbol{\sigma_{\mathrm{sgp}}}$} 
& \multirow{2}{*}{\textbf{Metrics}} 
& \multicolumn{3}{c}{\textbf{SAC-SGPC}} 
& \multicolumn{3}{c}{\textbf{TD3-SGPC}} \\
\cmidrule(lr){3-5} \cmidrule(lr){6-8}
& & SafeHopper & SafeWalker2d & SafePointGoal1 
& SafeHopper & SafeWalker2d & SafePointGoal1 \\
\midrule
\multirow{2}{*}{0.02}
& Reward & 980.4$\pm$300.8 & 2861.0$\pm$181.0 & 19.0$\pm$9.9
         & 838.9$\pm$\textbf{331.8} & 2754.8$\pm$279.3 & 23.3$\pm$2.9 \\
& Cost   & 4.1$\pm$20.3    & 5.6$\pm$10.9     & \textbf{6.5}$\pm$\textbf{13.6}
         & 9.9$\pm$14.5    & 13.8$\pm$9.7     & 13.0$\pm$17.1 \\
\cmidrule{2-8}
\multirow{2}{*}{0.1}
& Reward & \textbf{1120.8}$\pm$84.9 & 2831.3$\pm$\textbf{111.1} & \textbf{25.6}$\pm$\textbf{1.4}
         & 826.9$\pm$521.1 & \textbf{2850.7}$\pm$\textbf{175.4} & \textbf{23.3}$\pm$2.4 \\
& Cost   & 4.0$\pm$6.9     & 1.5$\pm$3.1      & 12.6$\pm$16.7
         & \textbf{0.7}$\pm$\textbf{1.3}     & \textbf{11.7}$\pm$\textbf{7.2}     & \textbf{9.6}$\pm$\textbf{12.6} \\
\cmidrule{2-8}
\multirow{2}{*}{0.5}
& Reward & 1029.9$\pm$\textbf{15.3} & \textbf{2873.0}$\pm$177.1 & 24.0$\pm$4.0
         & \textbf{882.7}$\pm$384.8 & 2772.6$\pm$245.5 & 23.3$\pm$\textbf{2.1} \\
& Cost   & \textbf{0.0}$\pm$\textbf{0.0}     & \textbf{0.1}$\pm$\textbf{0.3}      & 12.4$\pm$16.2
         & 6.6$\pm$15.2    & 14.0$\pm$12.8    & 12.1$\pm$15.0 \\
\bottomrule
\end{tabular}
\label{table_ablation_range}
\end{table*}

\begin{table*}[!tb]
\caption{Hyperparameter analysis of the SGP sample number $n$}
\centering
\begin{tabular}{clcccccc}
\toprule
\multirow{2}{*}{$\boldsymbol{n}$} 
& \multirow{2}{*}{\textbf{Metrics}} 
& \multicolumn{3}{c}{\textbf{SAC-SGPC}} 
& \multicolumn{3}{c}{\textbf{TD3-SGPC}} \\
\cmidrule(lr){3-5} \cmidrule(lr){6-8}
& & SafeHopper & SafeWalker2d & SafePointGoal1 
& SafeHopper & SafeWalker2d & SafePointGoal1 \\
\midrule
\multirow{2}{*}{16}
& Reward & \textbf{1120.8}$\pm$\textbf{84.9}  & 2831.3$\pm$\textbf{111.1}   & 19.8$\pm$10.1 
         & 826.9$\pm$521.1  & \textbf{2850.7}$\pm$175.4   & 23.4$\pm$\textbf{1.8} \\
& Cost   & 4.0$\pm$6.9      & 1.5$\pm$3.1        & \textbf{8.0}$\pm$\textbf{13.8}
         & \textbf{0.7}$\pm$\textbf{1.3}      & 11.7$\pm$7.2       & 10.2$\pm$14.0 \\
\cmidrule{2-8}
\multirow{2}{*}{64}
& Reward & 1078.7$\pm$152.4 & 2839.2$\pm$160.5   & \textbf{25.6}$\pm$\textbf{1.4}
         & 1119.0$\pm$495.7 & 2794.4$\pm$106.3 & 23.3$\pm$2.4 \\
& Cost   & 41.9$\pm$72.0    & 0.1$\pm$0.3        & 12.6$\pm$16.7
         & 10.9$\pm$20.8  & 21.8$\pm$15.4      & 9.6$\pm$12.6 \\
\cmidrule{2-8}
\multirow{2}{*}{256}
& Reward & 1064.7$\pm$107.3 & 2819.9$\pm$152.4   & 20.0$\pm$10.0
         & \textbf{1126.5}$\pm$\textbf{129.6} & 2770.7$\pm$207.0   & 22.8$\pm$2.5 \\
& Cost   & 18.9$\pm$43.6    & \textbf{0}$\pm$\textbf{0}            & 9.7$\pm$14.4
         & 14.7$\pm$25.5    & 11.1$\pm$10.9      & \textbf{9.5}$\pm$\textbf{12.4} \\
\cmidrule{2-8}
\multirow{2}{*}{1024}
& Reward & 1100.9$\pm$85.9  & \textbf{2840.0}$\pm$162.3   & 19.5$\pm$10.2
         & 1038.8$\pm$354.8 & 2812.8$\pm$\textbf{81.3}    & \textbf{23.6}$\pm$2.6 \\
& Cost   & \textbf{1.0}$\pm$\textbf{1.8}      & 12.7$\pm$17.6      & 10.3$\pm$17.0
         & 20.6$\pm$33.0    & \textbf{5.5}$\pm$\textbf{4.0}        & 12.6$\pm$16.6 \\
\bottomrule
\end{tabular}
\label{table_ablation_num}
\end{table*}

\section{Conclusion}
This work investigates the trade-off between optimality and safety (i.e., constraint feasibility) in off-policy safe RL. We introduce SGP and SGPC, and formulate a novel COP by integrating them into the standard safe RL framework. The resulting formulation enables the optimization process to proactively guide the actor policy toward the safe set, thereby improving constraint satisfaction and providing a theoretical foundation for the proposed SGPC-based safe RL methods. In this process, SGP serves as a nearby safer reference policy, while SGPC encourages the actor policy to approach this reference during training. Based on this formulation, we develop an SGPC-based primal-dual off-policy safe RL algorithm and instantiate it with two practical implementations, SAC-SGPC and TD3-SGPC. Experiments on multiple Safety-Gymnasium environments show that SGPC-based methods generally improve empirical training stability and reduce constraint violations and cost oscillations while achieving competitive or better cumulative rewards. Future work will extend SGPC to broader RL algorithms, more complex constraints, and real-world robotic applications.

\appendices  
\numberwithin{equation}{section}
\numberwithin{figure}{section}
\numberwithin{table}{section}

\section{Proofs}
\label{App:B}
\subsection{Proof of Lemma~\ref{feasible_2cop}}
\label{App:B1}
    Suppose a policy $\pi$ is feasible for \eqref{eq4.1-1}. Then
    \begin{align}
    \mathbb{E}_{s\sim p_s,a\sim\pi}
    \left[Q_c^{\pi}(s,a+\Delta a)\right]&\leq d,
    \label{eq:app-feasible-cost}\\
    \mathbb{E}_{s\sim p_s,a\sim\pi}
    \left[\|\Delta a\|_1
    \mid\Delta a=\pi_{sgp}^+(s,a)\right]&=0.
    \label{eq:app-feasible-sgpc}
    \end{align}
    Since the $\ell_1$ norm is non-negative, \eqref{eq:app-feasible-sgpc} implies $\|\Delta a\|_1=0$ almost surely under the joint distribution induced by $s\sim p_s$ and $a\sim\pi$. Hence, $\Delta a=\boldsymbol{0}$ almost surely. Substituting this result into \eqref{eq:app-feasible-cost} gives
    $$\mathbb{E}_{s\sim p_s,a\sim\pi}\left[Q_c^{\pi}(s,a)\right]\leq d.$$
    This is the constraint in the original COP \eqref{eq3.1-1}. Therefore, $\pi$ is also feasible for \eqref{eq3.1-1}.

\subsection{Proof of Lemma~\ref{equal_2cop}}
\label{App:B2}
    Since $\|\Delta a\|_1$ is non-negative, satisfying the SGPC implies
    \[
    \Delta a=\boldsymbol{0}
    \quad\text{almost surely}.
    \]
    Substituting $\Delta a=\boldsymbol{0}$ into the objective and cost expressions of \eqref{eq4.1-1} yields
    \begin{align*}
    \mathbb{E}_{s\sim p_s,a\sim\pi}
    \left[Q_r^{\pi}(s,a+\Delta a)\right]
    &=
    \mathbb{E}_{s\sim p_s,a\sim\pi}
    \left[Q_r^{\pi}(s,a)\right],\\
    \mathbb{E}_{s\sim p_s,a\sim\pi}
    \left[Q_c^{\pi}(s,a+\Delta a)\right]
    &=
    \mathbb{E}_{s\sim p_s,a\sim\pi}
    \left[Q_c^{\pi}(s,a)\right].
    \end{align*}
    Therefore, for any policy satisfying SGPC, the objective and cost expressions in the SGPC-based COP \eqref{eq4.1-1} coincide with those in the original COP \eqref{eq3.1-1}.

\subsection{Proof of Lemma~\ref{optimal_2cop}}
\label{App:B3}
    Let $\mathcal{F}_{\mathrm{sgpc}}$ and $\mathcal{F}_{\mathrm{orig}}$ denote the feasible sets of \eqref{eq4.1-1} and \eqref{eq3.1-1}, respectively. By Lemma~\ref{feasible_2cop},   $\mathcal{F}_{\mathrm{sgpc}}\subseteq\mathcal{F}_{\mathrm{orig}}$. Suppose that $\pi^*$ is optimal for \eqref{eq3.1-1} and $\pi^*\in\mathcal{F}_{\mathrm{sgpc}}$. For every     $\pi\in\mathcal{F}_{\mathrm{sgpc}}$, we also have $\pi\in\mathcal{F}_{\mathrm{orig}}$. By     Lemma~\ref{equal_2cop}, the objective expressions of the two COPs coincide for policies in     $\mathcal{F}_{\mathrm{sgpc}}$. Therefore, the optimality of $\pi^*$ over $\mathcal{F}_{\mathrm{orig}}$ implies its optimality over $\mathcal{F}_{\mathrm{sgpc}}$; that is, $\pi^*$ is also optimal for \eqref{eq4.1-1}.

\section{Detailed Experimental Setup}
\label{App:C}
\subsection{Environments}
We evaluate all methods on tasks from Safety-Gymnasium~\cite{Safety_Gymnasium_2023}. These environments cover two representative categories of safe RL tasks: velocity-constrained locomotion and safe navigation. 
\subsubsection{Locomotion Environments}
SafeAnt, SafeHalfCheetah, SafeHopper, SafeWalker2d, and SafeHumanoid are velocity-constrained locomotion tasks, where the agent is encouraged to maximize locomotion performance while satisfying velocity-related safety constraints. For these locomotion environments, at each time step, a binary cost of 1 is incurred if the agent's velocity exceeds the environment-specific threshold, i.e.,
$$cost = bool(V_{current}>V_{threshold}),$$ where the thresholds are listed in
Table~\ref{tab:velocity_thresholds}.
\begin{table}[tb]
\caption{Velocity thresholds for locomotion environments}
\centering
\small
\begin{tabular}{lc}
\toprule
Environment & Velocity threshold $\bar v_e$ \\
\midrule
SafeAnt         & 2.6222 \\
SafeHalfCheetah & 3.2096 \\
SafeHopper      & 0.7402 \\
SafeWalker2d    & 2.3415 \\
SafeHumanoid    & 1.4149 \\
\bottomrule
\end{tabular}
\label{tab:velocity_thresholds}
\end{table}

\subsubsection{Navigation Environments}
SafePointGoal1, SafeCarGoal1, and SafePointButton1 are navigation tasks in
which the agent must complete a goal-reaching or button-touching objective
while avoiding unsafe interactions. For all navigation tasks, the reward is
defined as
\begin{equation}
    r_t = \beta(d_{t-1}-d_t)
    + bool(\text{success at }t) R_{\mathrm{goal}},
\end{equation}
where $d_t$ is the distance to the current target, $\beta$ is a discount factor, and $R_{\mathrm{goal}}$ is the bonus for reaching
the goal or touching the target button. In SafePointGoal1 and SafeCarGoal1, the agent must navigate to the goal while avoiding eight hazards. One vase is also present
and observable but does not contribute to the cost. A binary cost is incurred
when the agent enters a hazard region. In SafePointButton1, the agent must
contact a designated target selected from four buttons while avoiding two
gremlins and non-target buttons. Contact with a gremlin or a non-target
button incurs a binary cost. For each navigation task, observable object
classes are represented by 90 LiDAR rays. For each observable object class, we use a 90-bin LiDAR observation. At reset, the orientations of the agent and objects are sampled uniformly from $[0,2\pi)$, and their locations are sampled uniformly from $[-1.5,1.5]^2$, subject to the environment's placement constraints. A new goal or target button is sampled after each success, and each episode contains
at most 1000 steps.

\subsection{Baselines}
\subsubsection{Off-policy baselines}
We compare SGPC-based methods with five representative off-policy safe RL baselines: SAC-Lag and TD3-Lag, two Lagrangian-based algorithms~\cite{Lag_RL-Ha_2021,Lag_RL_Ray_2019}; SAC-PID and TD3-PID, two PID Lagrangian-based algorithms~\cite{PIDLag_2020}; and CVPO~\cite{CVPO}. SAC-Lag, TD3-Lag, SAC-PID, and TD3-PID are implemented using the OmniSafe codebase~\cite{omnisafe_2024} (\url{https://github.com/PKU-Alignment/omnisafe}). Unless otherwise specified, we use the default hyperparameters provided by the corresponding OmniSafe implementations. For CVPO, we use its official implementation from the FSRL codebase~\cite{fsrl_cvpo} (\url{https://github.com/liuzuxin/fsrl}). Because the original CVPO experiments used different environment variants that differ from the official Safety-Gymnasium environments, we align the maximum episode length, total number of environment steps, and update frequency with our experimental protocol to ensure a fair comparison across algorithms.

\subsubsection{On-policy baselines}
For additional comparison, we also evaluate two representative on-policy safe RL methods, CPO~\cite{CPO_2017} and CUP~\cite{CUP_2022}. CPO and CUP are also implemented using the OmniSafe codebase~\cite{omnisafe_2024} (\url{https://github.com/PKU-Alignment/omnisafe}).

\subsection{Implementation Details and Hyperparameter Settings}
\begin{table}[t]
\caption{Default hyperparameters of SGPC-based methods}
\centering
\renewcommand{\arraystretch}{0.7}
\begin{tabular}{lcc}
\toprule
\textbf{Hyperparameter} 
& \tabincell{c}{\textbf{Velocity}\\\textbf{Tasks}} 
& \tabincell{c}{\textbf{Navigation}\\\textbf{Tasks}} \\
\midrule
Replay buffer size & $10^6$ & $10^6$ \\
Batch size & 256 & 256 \\
Discount factor $\gamma$ & 0.99 & 0.99 \\
Soft update coefficient $\tau$ & 0.005 & 0.005 \\
Actor learning rate & $3\times10^{-4}$ & $5\times10^{-6}$ \\
Critic learning rate & $3\times10^{-4}$ & $10^{-3}$ \\
Hidden sizes & $(256,256)$ & $(256,256)$ \\
Activation & ReLU & ReLU \\
Init. multiplier $\lambda$ & $10^{-3}$ & 0 \\
Init. multiplier $\mu$ & 0 & 0 \\
Learning rate $\eta_\lambda$ & $10^{-5}$ & $5\times10^{-7}$ \\
Learning rate $\eta_\mu$ & $10^{-4}$ & $10^{-6}$ \\
SGP sample number $n$ & 16 & 64 \\
SGP sampling range $\sigma_{\mathrm{sgp}}$ & 0.1 & 0.1 \\
Penalty coefficient $\lambda_{\mathrm{sgp}}$ & 0.1 & 0.1 \\
Primal-dual ratio $\rho$ & 0.5 & 0.5 \\
\bottomrule
\end{tabular}
\label{tab:sgpc_hyperparams}
\end{table}
We implement SAC-SGPC and TD3-SGPC by incorporating SGP and SGPC into the actor update process of SAC-Lag and TD3-Lag in the OmniSafe codebase~\cite{omnisafe_2024}. Except for SGPC-specific hyperparameters, all network architectures and training-related hyperparameters are kept identical to the corresponding Lagrangian-based implementations in OmniSafe. For all algorithms and environments, the maximum episode length is fixed to 1000. The total number of environment steps is set to $1\times10^6$ for SafeAnt, SafeHalfCheetah, SafeHopper, and SafeWalker2d, and $3\times10^6$ for SafeHumanoid, SafePointGoal1, SafeCarGoal1, and SafePointButton1. The default SGPC-specific hyperparameters are listed in Table~\ref{tab:sgpc_hyperparams}; unless otherwise specified, all experiments of the proposed algorithms are conducted using these default hyperparameter settings.

\bibliographystyle{IEEEtran}
\bibliography{reference}

\vfill

\end{document}